\pdfoutput=1

\documentclass[11pt]{article}

\usepackage{ACL2023}

\usepackage{times}
\usepackage{latexsym}

\usepackage[T1]{fontenc}

\usepackage[utf8]{inputenc}

\usepackage{microtype}
\usepackage{inconsolata}
\usepackage{helvet}
\usepackage{courier}
\usepackage{stfloats}
\usepackage{color}
\usepackage{amsmath}
\usepackage{amssymb}
\usepackage{array}
\usepackage{graphicx}
\usepackage{subfigure}
\usepackage{colortbl} 
\usepackage{arydshln} 
\usepackage{multirow} 
\usepackage{multicol}
\usepackage[encapsulated]{CJK}
\usepackage{float}
\usepackage{mathrsfs}
\usepackage{enumitem}
\usepackage{mathtools}
\usepackage{amsopn}
\usepackage{url}
\usepackage{bm}
\usepackage{soul}
\usepackage{pgfplots}
\usepackage{booktabs}
\usepackage{makecell}
\usepackage{algorithm}
\usepackage{algorithmic}
\usepackage{enumitem}
\usepackage{amsthm}
\usepackage{ulem}
\pgfplotsset{compat=1.14}
\newtheorem{proposition}{Proposition}

\usepackage{CJKutf8}
\usepackage{tablefootnote}
\usepackage[para,online,flushleft]{threeparttable}
\usepackage{ascii}

\title{Rethinking Translation Memory Augmented Neural Machine Translation}

\author{
 Hongkun Hao$^1$\thanks{\ \ Partial work was done when Hongkun Hao was interning at Tencent AI Lab.}  \quad Guoping Huang$^2$ \quad Lemao Liu$^{2}$\footnotemark[2]  \\ 
 \bf Zhirui Zhang$^{2}$  \quad  Shuming Shi$^2$ \quad Rui Wang$^{1}$\thanks{\ \ Lemao Liu and Rui Wang are corresponding authors.} \\
$^1$Shanghai Jiao Tong University \qquad { \normalsize \{haohongkun, wangrui12\}@sjtu.edu.cn} \\
$^2$Tencent AI Lab \qquad  { \normalsize \{donkeyhuang, redmondliu, shumingshi\}@tencent.com} \quad { \normalsize zrustc11@gmail.com} \\
}

\begin{document}
\maketitle
\begin{abstract}
This paper rethinks translation memory augmented neural machine translation (TM-augmented NMT) from two perspectives, i.e., a probabilistic view of retrieval and the variance-bias decomposition principle. The finding demonstrates that TM-augmented NMT is good at the ability of fitting data (i.e., lower bias) but is more sensitive to the fluctuations in the training data (i.e., higher variance), which provides an explanation to a recently reported contradictory phenomenon on the same translation task: TM-augmented NMT substantially advances vanilla NMT under the high-resource scenario whereas it fails under the low-resource scenario. Then we propose a simple yet effective TM-augmented NMT model to promote the variance and address the contradictory phenomenon. Extensive experiments show that the proposed TM-augmented NMT achieves consistent gains over both conventional NMT and existing TM-augmented NMT under two variance-preferable (low-resource and plug-and-play) scenarios as well as the high-resource scenario. 
\end{abstract}

\begin{CJK*}{UTF8}{gbsn}

\section{Introduction}
\label{section:introduction}

The effectiveness of Translation Memory (TM) in Machine Translation has long been recognized~\cite{garcia2009beyond, koehn2010convergence,utiyama2011searching,wang2013integrating, liu2019unified}, because a TM retrieved from a bilingual dataset (i.e., training data or an external dataset) may provide valuable knowledge for the source sentence to be translated. Many notable approaches recently have been proposed to enhance neural machine translation (NMT) by using a TM \citep{feng-etal-2017-memory,gu2018search,DBLP:conf/ecai/CaoKX20,hoang2022improve_Robustness_TM_NMT,cai-etal-2021-neural,2105.13072}. 


For example, on the standard JRC-Acquis task, TM-augmented NMT achieves substantial gains over the vanilla NMT (without TM) under the conventional high-resource training scenario. Unfortunately, \citet{cai-etal-2021-neural} surprisingly find that TM-augmented NMT fails to advance the vanilla NMT model 
on the same task under a low-resource scenario, if a quarter of the full data is used for training and TM retrieval, 
as reproduced in Table ~\ref{table:preliminary}. Due to the lack of theoretical insights, it is unclear the reason why such a contradictory phenomenon happens. This motivates us to rethink the working mechanism of TM-augmented NMT as well as its statistical principle. 

\begin{table}[t]
\centering
\begin{tabular}{l|l|l}
\hline
\textbf{Model} & \textbf{High-Resource} & \textbf{Low-Resource} \\
\hline
\makecell[l]{w/o TM} &  $60.83$  & $54.54$   \\
\hline
\makecell[l]{w/ TM} &  $63.76\uparrow$  & $53.92\downarrow$  \\
\hline
\end{tabular}
\caption{Testing BLEU comparison on JRC-Acquis German$\Rightarrow$English task. w/o TM and w/ TM denote the vanilla Transformer and TM-augmented Transformer, respectively; High-Resource and Low-Resource denote full and quarter train data are used for NMT training and TM retrieval.
}
\label{table:preliminary}
\end{table}

In this paper, we first cast TM-augmented NMT as an approximation of a latent variable model where the retrieved TM is the latent variable through a probabilistic view of retrieval. From this statistical viewpoint, we identify that the success of such an approximation depends on the variance of TM-augmented NMT with respect to the latent variable. Then, we empirically estimate the variance of TM-augmented NMT from the principle of variance-bias decomposition in the learning theory. Our findings demonstrate that TM-augmented NMT is worse than the vanilla NMT in terms of variance which indicates the sensitivity to fluctuations in the training
set, although TM-augmented NMT is better in terms of the bias which indicates the ability of fitting data. The finding about the variance takes the responsibility of the contradictory phenomenon in Table ~\ref{table:preliminary}, because limited training data may amplify its negative effect on variance~\citep{vapnik1999overview,niyogi1996generalization,bishop2006pattern}. 

To better trade off the variance and the bias, we further propose a simple yet effective method for TM-augmented NMT. The proposed method is general to be applied on top of any TM-augmented NMT models. To validate the effectiveness of the proposed approach, we conduct extensive experiments on several translation tasks under different scenarios including low-resource scenario, plug-and-play scenario, and high-resource scenario. 

Contributions of this paper are three-fold:
\begin{itemize}[wide=0\parindent,noitemsep, topsep=0pt]
    \item It rethinks and analyzes the variance of TM-augmented NMT models from the probabilistic view of retrieval and the bias-variance decomposition perspective.
    \item It proposes a simple yet effective lightweight network to ensemble TM-augmented NMT models, which better trades off variance and bias.
    \item Its experiments show the effectiveness of the aforementioned approach, which outperforms both vanilla Transformer and baseline TM-augmented NMT models under the low-resource scenario, plug-and-play scenario, and conventional high-resource scenario.
\end{itemize}

\section{Preliminary}
\label{section:preliminary}

\subsection{NMT}
Suppose $\mathbf{x}=\{x_1, \cdots, x_n\}$ is a source sentence and $\mathbf{y} = \{y_1, \cdots, y_m\}$ is the corresponding target sentence. NMT builds a probabilistic model with neural networks parameterized by $\boldsymbol{\theta}$, which is used to translate $\mathbf{x}$ in the source language to $\mathbf{y}$ in the target language. Formally, NMT aims to generate output $\mathbf{y}$ given $\mathbf{x}$ according to the conditional probability defined by Eq.~\eqref{equation:NMT_vanilla}:
\begin{equation}
\label{equation:NMT_vanilla}
\begin{aligned}
        P(\mathbf{y}|\mathbf{x}; \boldsymbol{\theta)} & = \prod_{t=1}^m P(y_t | \mathbf{x}, \mathbf{y}_{<t}; \boldsymbol{\theta}) \\
        & = \prod_{t=1}^m \text{Softmax}\big( f(H_t) \big)[y_t] ,
\end{aligned}
\end{equation}
\noindent where $H_t$ denotes the NMT decoding state. 

\subsection{TM-Augmented NMT}
In general, TM-augmented NMT works in the following two-step paradigm.  
It first retrieves top-$K$ TM bilingual sentences $\mathbf{Z} = \{\mathbf{z_k}\}_{k=1}^K$, where $\mathbf{z}_k = ( \mathbf{x}_k^\text{tm}, \mathbf{y}_k^\text{tm} )$ is the $k$-th TM; Then it generates the translation $\mathbf{y}$ by using the information from the source sentence $\mathbf{x}$ and its retrieved TMs $\mathbf{Z}$. 

\paragraph{Retrieval Model}
Following previous works \citep{gu2018search,zhang-etal-2018-guiding,Xia_Huang_Liu_Shi_2019,he-etal-2021-fast}, for $\mathbf{x}$ we employ Apache Lucene \citep{bialecki2012apache} to retrieve top-100 similar bilingual sentences from datastore. 
Then we adopt the similarity function in Eq.~\eqref{equation:retrieval} to re-rank the retrieved bilingual sentences and maintain top-$K$ (e.g. $K=5$) bilingual sentences as the TMs for $\mathbf{x}$:
\begin{equation}
\label{equation:retrieval}
    \text{sim}(\mathbf{x}, \mathbf{z}_k) = 1 - \frac{\text{dist}(\mathbf{x}, \mathbf{x}_k^\text{tm})}{\max(|\mathbf{x}|, |\mathbf{x}_k^\text{tm}|) },
\end{equation}
where $\text{dist}$ denotes the edit-distance.

\paragraph{Generation Model}
Given a source sentence $\mathbf{x}$ and a small set of relevant TMs $\mathbf{Z} = \{\mathbf{z_k}\}_{k=1}^K$, the generation model defines the conditional probability $P(\mathbf{y}|\mathbf{x}, \mathbf{Z} ; \boldsymbol{\theta})$:
\begin{equation}
\label{equation:TM-nmt-original}
\begin{aligned}
        P(\mathbf{y}|\mathbf{x}, \mathbf{Z} ; \boldsymbol{\theta}) & =    \prod_{t=1}^m P(y_t | \mathbf{x}, \mathbf{y}_{<t}, \mathbf{Z}; \boldsymbol{\theta}) \\
        & =    \prod_{t=1}^m \text{Softmax}\big( f(H_{t, \mathbf{Z}}) \big)[y_t],
\end{aligned}
\end{equation}
where $H_{t, \mathbf{Z}}$ denotes the decoding state of TM-augmented NMT. There are different TM-augmented NMT models, and accordingly there are different instantiations of $H_{t, \mathbf{Z}}$. We refer readers to ~\citet{gu2018search,bulte-tezcan-2019-neural,cai-etal-2021-neural,he-etal-2021-fast} for its detailed definitions.

\section{Rethinking TM-Augmented NMT}
\subsection{Probabilistic View of Retrieval}
\label{subsection:view_from_latent_variable_model}
Given the source sentence $\mathbf{x}$, the top-$K$ retrieval aforementioned actually can be considered as a probabilistic retrieval model (i.e., $P(\mathbf{Z} | \mathbf{x})$), from which a translation memory $\mathbf{Z} = \{\mathbf{z}_k\}_{k=1}^K$ is sampled.
Mathematically, such a retrieval model $P(\mathbf{Z} | \mathbf{x})$ is defined as follows:

\begin{equation}
    P(\mathbf{Z}|\mathbf{x})=\prod_k P(\mathbf{z}_k|\mathbf{x}) \propto \exp (\text{sim}(\mathbf{x}, \mathbf{z}_k)/T) ,
\end{equation}
\noindent where $\text{sim}$ is defined as in Eq.~\eqref{equation:retrieval}, and $T>0$ is a temperature. Note that if $T$ is a sufficiently small number, sampling $\mathbf{z}_k$ from the above probabilistic retrieval model is similar to the deterministic $\arg\max$ retrieval widely used in prior studies~\citep{gu2018search, zhang-etal-2018-guiding,Xia_Huang_Liu_Shi_2019}.

By using the probabilistic retrieval model $P(\mathbf{Z} | \mathbf{x})$, the translation model $P(\mathbf{y}|\mathbf{x})$ is related to the variable $\mathbf{Z}$ theoretically through the following latent variable model:
\begin{equation}
\label{eq:lvm}
    P(\mathbf{y}|\mathbf{x}) =  \sum_\mathbf{Z} P(\mathbf{Z} | \mathbf{x}) P(\mathbf{y}|\mathbf{x}, \mathbf{Z}) = \mathbb{E}_\mathbf{Z}P(\mathbf{y}|\mathbf{x},\mathbf{Z}) .
\end{equation}
In practice, it is impossible to perform the summation over all possible $\mathbf{Z}$. Instead it can be estimated by the Monte Carlo sampling:
\begin{equation}
\label{eq:TM-lvm}
    P(\mathbf{y}|\mathbf{x}) \approx {P(\mathbf{y} | \mathbf{x}, \mathbf{Z})}, \text{ with } \mathbf{Z}\sim P(\mathbf{Z} | \mathbf{x})  .
\end{equation}
As a result, according to Eq.~\eqref{eq:TM-lvm}, we can see 
the following statement: {the TM-augmented NMT model $P(\mathbf{y} | \mathbf{x}, \mathbf{Z})$ can be considered as an approximation of $P(\mathbf{y} | \mathbf{x})$ via Monte Carlo sampling over a latent variable model in Eq.~\eqref{eq:lvm}.} In particular, whether TM-augmented NMT $P(\mathbf{y} | \mathbf{x}, \mathbf{Z})$ is a good estimator depends on the {\bf expected approximate error} defined by $\mathbb{E}_\mathbf{Z}\big ( P(\mathbf{y}|\mathbf{x},\mathbf{Z}) -P(\mathbf{y}|\mathbf{x})\big)^2$.  In other words, $P(\mathbf{y}|\mathbf{x},\mathbf{Z})$ is a good estimator of $P(\mathbf{y}|\mathbf{x})$ if the expected approximate error is small; 
otherwise $P(\mathbf{y}|\mathbf{x},\mathbf{Z})$ is not a good estimator \citep{voinov2012unbiased}. 



Because of the Equation~\eqref{eq:lvm}, the expected estimation error is actually derived by the variance of $P(\mathbf{y}|\mathbf{x},\mathbf{Z})$ with respect to $\mathbf{Z}$ as follows:
\begin{equation}
\begin{aligned}
    & \mathbb{E}_\mathbf{Z}\big ( P(\mathbf{y}|\mathbf{x},\mathbf{Z}) -P(\mathbf{y}|\mathbf{x})\big)^2 =  \\
     & \mathbb{E}_\mathbf{Z}\big ( P(\mathbf{y}|\mathbf{x},\mathbf{Z}) -\mathbb{E}_\mathbf{Z}P(\mathbf{y}|\mathbf{x},\mathbf{Z})\big)^2 
     := \mathbb{V}_\mathbf{Z}P(\mathbf{y}|\mathbf{x},\mathbf{Z})
.
\end{aligned}
\end{equation}
From the above equation, it is easy to observe that $\mathbb{V}_\mathbf{Z}P(\mathbf{y}|\mathbf{x},\mathbf{Z})$ actually controls the approximate effect of $P(\mathbf{y}|\mathbf{x},\mathbf{Z})$.
Therefore, $\mathbb{V}_\mathbf{Z}P(\mathbf{y}|\mathbf{x},\mathbf{Z})$ would have a negative effect on TM-augmented NMT due to the fluctuations with respect to the variable $\mathbf{Z}$ if $\mathbb{V}_\mathbf{Z}P(\mathbf{y}|\mathbf{x},\mathbf{Z})$ is relatively large.

It is worth noting that the above analysis on the variance is only related to the variable $\mathbf{Z}$, but has nothing to do with the variables $\mathbf{x}$ and $\mathbf{y}$ and it is agnostic to neural network architecture. Moreover, it is intractable to estimate the approximate error, i.e., the variance $\mathbb{V}_\mathbf{Z}P(\mathbf{y}|\mathbf{x},\mathbf{Z})$ with respect to $\mathbf{Z}$ because $P(\mathbf{y}|\mathbf{x})$ requires the summation over all possible $\mathbf{Z}$. 
In the next subsection, we will employ the variance-bias decomposition principle to quantify the variance with respect to all variables given specific models.

\subsection{Variance-Bias Decomposition}
\label{subsection:bias_variance_decomposition}
\paragraph{Estimation of Bias and Variance}
Bias-variance trade-off is a fundamental principle for understanding the generalization of predictive learning models and larger variance may induce the lower generalization ability of models \citep{geman1992neural, hastie2009elements, yang2020rethinking}. 

The bias-variance decomposition is typically defined in terms of the Mean Squared Error at the example level for classification tasks and~\citet{yang2020rethinking} reorganize its definition in terms of the Cross-Entropy loss. 
In the machine translation task the optimization loss is the Cross-Entropy loss at the token level, we hence simply extend the variance and bias decomposition in terms of cross-entropy~\citep{yang2020rethinking} to the token level.\footnote{Since it is not clear how to extend the variance-bias decomposition on top of the BLEU score used as an evaluation metric, we instead consider it on top of the token-level cross-entropy loss.}
Specifically, assume $P_0(y|\mathbf{x}, \mathbf{y}_{<t})$ is the empirical distribution (i.e., $P_0(y|\mathbf{x}, \mathbf{y}_{<t})$ is one if $y=y_t$ is the ground-truth word $y_t$ and 0 otherwise), and $P(y|\mathbf{x}, \mathbf{y}_{<t})$ is the model output distribution. Then the expected cross entropy with respect to a random variable $\tau=\{\langle \mathbf{x} , \mathbf{y}\rangle\}$ (i.e., the training data) can be defined: 
\begin{multline}
    \mathbb{E}_{\tau} \big(H(P_0, P)\big) = \\ -\mathbb{E}_{\tau} \big(\sum_t \sum\limits_{y} P_0(y |\mathbf{x}, \mathbf{y}_{<t}) \log P(y|\mathbf{x}, \mathbf{y}_{<t}) \big)  .
\end{multline}
Further, it can be decomposed as:
\begin{multline}
 \label{equation:decomposition}
    \mathbb{E}_{\tau}\big(H(P_0, P)\big) = \underbrace{D_\text{KL}(P_0 || \overline{P})}_{\textbf{Bias}^2}  \\ +
    \underbrace{\mathbb{E}_{\tau}\big(D_\text{KL}(P || \overline{P})\big)}_{\textbf{Variance}} ,
\end{multline}
\noindent
where $\overline{P}$ is the expected probability after normalization:
\begin{equation}
\overline{P}(y|\mathbf{x}, \mathbf{y}_{<t}) \propto \exp\Big(\mathbb{E}_{\tau}\big( \log {P}(y|\mathbf{x}, \mathbf{y}_{<t})\big) \Big).
\end{equation}
Generally speaking, the bias indicates the ability of the model $P$ to fit the data whereas 
the variance measures the sensitivity of the model $P$ to fluctuations in the training
data. 

We follow classic methods \citep{hastie2009elements, yang2020rethinking} to estimate the variance in Eq.~\eqref{equation:decomposition}, which is shown in Algorithm \ref{algorithm:estimate_variance} of Appendix \ref{appendix_sec:variance_estimation_algorithm}. The key idea is to estimate the expectation over the random variable $\tau$ and it can be achieved by randomly splitting the given training dataset into several parts and training several models on each part for average. 
The above bias and variance estimation method is defined for $P(y | \mathbf{x}, \mathbf{y}_{<t})$ but it is similar to estimate the bias and variance for the TM-augmented NMT $P(y|\mathbf{x}, \mathbf{y}_{<t}, \mathbf{Z})$ by retrieving a TM $\mathbf{Z}$ for each $\mathbf{x}$ via top-K retrieval as default rather than sampling as in \S\ref{subsection:view_from_latent_variable_model}.  

\paragraph{Experiments}
We conduct experiments on JRC-Acquis German$\Rightarrow$English task to estimate variance and bias of vanilla Transformer and TM-augmented NMT models. 
Similar to the preliminary experiment in \S\ref{section:introduction}, we retrieve top-5 TMs for TM-augmented NMT.
In order to eliminate the effect of model architecture, we use both TM-augmented NMT backbones, which respectively involve a single encoder~\citep{bulte-tezcan-2019-neural} and dual encoder~\citep{cai-etal-2021-neural}, see details in Appendix \ref{appendix_sec:backbone}.
Following \citet{yang2020rethinking}, $\text{Bias}^2$ is estimated by subtracting the variance from the loss. 

\begin{table}[t]
\small
\renewcommand\arraystretch{2}
\centering
\begin{tabular}{c|c|c|c|c}
\hline
\multirow{2}{*}{\textbf{Model}} & \multicolumn{2}{c|}{\textbf{Single Encoder}} & \multicolumn{2}{c}{\textbf{Dual Encoder}} \\
\cline{2-5}
& \textbf{Var} & \textbf{$\text{Bias}^2$} & \textbf{Var} & \textbf{$\text{Bias}^2$} \\
\hline
\makecell[c]{w/o TM} &  $\mathbf{0.2088}$ & 1.9519 &  $\mathbf{0.1573}$ & 1.9992  \\
\hline
\makecell[c]{w/ TM} &  0.2263 & $\mathbf{1.7500}$ & 0.2168 & $\mathbf{1.8460}$  \\
\hline
\end{tabular}
\caption{Estimated variance and bias on JRC-Acquis German$\Rightarrow$English task under single encoder and dual encoder backbone respectively. 
}
\label{table:variance_two_backbone}
\end{table}

Table ~\ref{table:variance_two_backbone} shows the variance-bias decomposition results of different models. 
Firstly, within each backbone, we can find that the variance of TM-augmented NMT model is larger than that of vanilla Transformer, which verifies our hypothesis that variance of TM-augmented NMT model is worse than that of vanilla Transformer, which results in poor performance under the low-resource scenario indicated in \S\ref{section:introduction}. 
Secondly, the bias of TM-augmented NMT model within each backbone is smaller than that of vanilla Transformer, which explains the better performance under the high-resource scenario indicated in \S\ref{section:introduction}. In addition, since the variance is highly dependent on the scale of training data and the limited training data may even amplify its negative effect on variance~\citep{niyogi1996generalization,vapnik1999overview,bishop2006pattern}, the higher variance of TM-augmented NMT in Table~\ref{table:variance_two_backbone} gives an explanation for the contradictory phenomenon in Table~\ref{table:preliminary}.  

In summary, although the TM-augmented NMT model has lower bias, resulting in fitting on the training data better especially when the training data size is large, it has non-negligible higher variance, which leads to its poor performance with less training data. Therefore, the inherent flaw drives us to find ways to reduce the variance of TM-augmented NMT models, as shown in the next section. Note that our proposed methods shown below can all theoretically reduce the variance with respect to the variable $\mathbf{Z}$ (i.e. TM) in \S \ref{subsection:view_from_latent_variable_model}, and our experiments below empirically show that reducing the variance with respect to $\mathbf{Z}$ can actually reduce the model variance introduced here.

\section{Proposed Approach for Lower Variance}
\label{section:proposed_approach}
In this section, we first propose two techniques to reduce the variance of the TM-augmented NMT model. Then we propose a new TM-augmented NMT on the basis of the two techniques to address the contradictory phenomenon as presented in \S\ref{section:introduction}.  
Finally we empirically quantify the variance and bias for the proposed TM-augmented NMT. Note that the proposed TM-augmented NMT is general enough to be applied on top of any specific TM-augmented NMT models. 

\subsection{Two Techniques to Reduce Variance}

\paragraph{Technique 1: Conditioning on One TM Sentence}
\label{subsection:change_number_of_TMs}
In conventional TM-augmented NMT, the translation model $P(\mathbf{y}|\mathbf{x}, \mathbf{Z})$ is conditioned on $\mathbf{Z}$ consisting of five\footnote{Since many studies \citep{gu2018search, Xia_Huang_Liu_Shi_2019, cai-etal-2021-neural} find that using five TMs is suitable.} retrieved TMs $\{\mathbf{z}_1, \mathbf{z}_2, \mathbf{z}_3, \mathbf{z}_4, \mathbf{z}_5\}$ with $\mathbf{z}_i = ( \mathbf{x}_i^\text{tm}, \mathbf{y}_i^\text{tm} )$. 
Similar to the analysis in Eq.(\ref{eq:lvm}-\ref{eq:TM-lvm}), we can obtain the following equations:
\begin{equation}
\begin{aligned}
    & P(\mathbf{y}|\mathbf{x}, \mathbf{z}_1) \\
    = & \sum_{\mathbf{z}_2, \cdots, \mathbf{z}_5} P(\mathbf{y}|\mathbf{x}, \mathbf{z}_1, \mathbf{Z}_{>1}) P(\mathbf{Z}_{>1}|\mathbf{x}, \mathbf{z}_1)  \\
  = & \sum_{\mathbf{z}_2, \cdots, \mathbf{z}_5} P(\mathbf{y}|\mathbf{x}, \mathbf{z}_1, \mathbf{Z}_{>1}) P(\mathbf{Z}_{>1}|\mathbf{x})   \\
  \approx & P(\mathbf{y}|\mathbf{x}, \mathbf{Z}), \text{  with  } \mathbf{Z}_{>1} \sim P(\mathbf{Z}_{>1}|\mathbf{x})  , 
\end{aligned}
\end{equation}
\noindent where $\mathbf{Z}_{>1}=\{\mathbf{z}_2, \mathbf{z}_3, \mathbf{z}_4, \mathbf{z}_5\}$, and the second equation holds due to the conditional independence assumption in Eq.~\eqref{eq:TM-lvm}. In this sense, we can see that $P(\mathbf{y}|\mathbf{x}, \mathbf{Z})$ conditioned on five retrieved TMs is actually an approximation of $P(\mathbf{y}|\mathbf{x}, \mathbf{z}_1)$ conditioned on a single retrieved TM $\mathbf{z}_1$.  As a result, as analysed in \S\ref{subsection:view_from_latent_variable_model}, whether $P(\mathbf{y}|\mathbf{x}, \mathbf{Z})$ is a good estimator depends on the variance with respect to $\mathbf{Z}_{>1}$, i.e., $\mathbb{V}_{\mathbf{Z}_{>1}} P(\mathbf{y}|\mathbf{x}, \mathbf{z}_1, \mathbf{Z}_{>1})$.




Based on the above analyses, to alleviate the variance, we directly estimate $P(\mathbf{y}|\mathbf{x})$ by using a single sampled sentence pair. Formally, 
suppose $\mathbf{z}=\langle \mathbf{x}^\text{tm}, \mathbf{y}^\text{tm} \rangle $ is a {\bf single} bilingual sentence sampled from $P(\mathbf{z}|\mathbf{x})$, we adopt the following TM-augmented NMT conditioned on a single retrieved sentence $\mathbf{z}$:
\begin{equation}
\label{eq:tm-cond-one}
\begin{aligned}
    P(y_t | \mathbf{x}, \mathbf{y}_{<t}) & =\mathbb{E}_z P(y_t | \mathbf{x}, \mathbf{y}_{<t}, \mathbf{z}) \\
    & \approx P(y_t | \mathbf{x}, \mathbf{y}_{<t}, \mathbf{z}) ,\\
    P(\mathbf{y}|\mathbf{x}) & = \prod_{t} P(y_t | \mathbf{x}, \mathbf{y}_{<t}) \\
    & \approx \prod_{t} P(y_t | \mathbf{x}, \mathbf{y}_{<t}, \mathbf{z}) , 
\end{aligned}
\end{equation}
\noindent where $P(y_t | \mathbf{x}, \mathbf{y}_{<t}, \mathbf{z})$ can be any model architecture of TM-augmented NMT models $P(y_t | \mathbf{x}, \mathbf{y}_{<t}, \mathbf{Z})$ by replacing top-$K$ TMs $\mathbf{Z}$ with a single top-1 TM $\mathbf{z}$. In addition, the training of the above model $P(y_t | \mathbf{x}, \mathbf{y}_{<t}, \mathbf{z})$ is the same as the training of the conventional model $P(y_t | \mathbf{x}, \mathbf{y}_{<t}, \mathbf{Z})$.

\paragraph{Technique 2: Enlarging the Sample Size}
\label{subsection:average_ensemble}
In Eq.~\eqref{eq:tm-cond-one}, $\mathbb{E}_z P(y_t | \mathbf{x}, \mathbf{y}_{<t}, \mathbf{z})$ is approximated by one sample $\mathbf{z}$, which still induces some potential estimation errors due to the variance $\mathbb{V}_\mathbf{z} P(y_t | \mathbf{x}, \mathbf{y}_{<t}, \mathbf{z})$. In fact, the estimation errors can be further reduced by the estimation using multiple samples as follows.
\begin{proposition}
\label{proposition:variance}
If $\mathbf{z}_1, \cdots, \mathbf{z}_K$ are independent and identically distributed random variables sampled from the $P(\mathbf{z}|\mathbf{x})$, then the following inequality holds (The proof is presented in Appendix \ref{appendix_sec:proof_of_lemma}.):
\begin{multline}
    \mathbb{V}_{\mathbf{z}} P(y_t | \mathbf{x}, \mathbf{y}_{<t}, \mathbf{z}) \ge \\ \mathbb{V}_{\mathbf{z}_1, \cdots, \mathbf{z}_K} \Big(\frac{1}{K}\sum_k P(y_t | \mathbf{x}, \mathbf{y}_{<t}, \mathbf{z}_k)\Big)  .
\end{multline}
\end{proposition}



According to the above Proposition, we propose the method to approximate $P(y_t|\mathbf{x}, \mathbf{y}_{<t})$ through the average ensemble of all $P(y_t | \mathbf{x}, \mathbf{y}_{<t}, \mathbf{z}_k)$:
\begin{equation}
\label{eq:tm-sum-k}
\begin{aligned}
& P(y_t | \mathbf{x}, \mathbf{y}_{<t}) \approx \frac{1}{K}\sum_{k=1}^K P(y_t | \mathbf{x}, \mathbf{y}_{<t}, \mathbf{z}_k) , \\
    & P(\mathbf{y}|\mathbf{x})  \approx  \prod_{t=1}^m \sum\limits_{k=1}^{K} P(y_t | \mathbf{x}, \mathbf{y}_{<t}, \mathbf{z}_{k}).
\end{aligned}
\end{equation}
Note that we do not particularly retrain the averaged ensemble model by directly taking the parameters from the model $P(y_t | \mathbf{x}, \mathbf{y}_{<t}, \mathbf{z})$ trained in Technique 1 above in this paper.


\subsection{TM-Augmented NMT via Weighted Ensemble}
\label{sub:weighted}

In our experiments as shown in \S\ref{subsection:empirical_analysis_variance} later, we found that the both techniques deliver better variance but sacrifice its ability of fitting data (i.e., bias) compared to the standard TM-augmented NMT conditioning on all retrieved $\mathbf{Z}=\{\mathbf{z}_k\}_{k=1}^K$.  
In order to further improve the bias for a better ability to fit data, we propose the weighted ensemble to establish a stronger relationship between the source sentence and each TM $\mathbf{z}_k$ by endowing a more powerful representation ability via the weighting coefficient $w(\mathbf{x}, \mathbf{y}_{<t}, \mathbf{z}_{k})$ in Eq.~\eqref{equation:TM-nmt-latend_variable_token_level}. Formally, the weighted ensemble model is defined as:
\begin{equation}
\label{equation:TM-nmt-latend_variable_token_level}
\begin{aligned}
    P(\mathbf{y}|\mathbf{x})  \approx  \prod_{t=1}^m \sum\limits_{k=1}^{K} w(\mathbf{x}, \mathbf{y}_{<t}, \mathbf{z}_{k}) P(y_t | \mathbf{x}, \mathbf{y}_{<t}, \mathbf{z}_{k}),
\end{aligned}
\end{equation}
\noindent where $P(y_t | \mathbf{x}, \mathbf{y}_{<t}, \mathbf{z}_{k})$ can be any TM-augmented model architecture as claimed in \S\ref{subsection:change_number_of_TMs}, and $w(\mathbf{x}, \mathbf{y}_{<t}, \mathbf{z}_{k})$ is a weighting network via the following equation:
\begin{equation}
\label{equation:w_k_in_TMNMT}
    w(\mathbf{x}, \mathbf{y}_{<t}, \mathbf{z}_{k}) = \text{Softmax}\big(f(H_t, {H_{t,k}})\big)[k],
\end{equation}
\noindent where $f$ consists of two linear layers with residual connection and layer normalization \citep{vaswani2017attention}, $H_t$ is the decoding state of a vanilla translation model $P(y_t|\mathbf{x}, \mathbf{y}_{<t})$, $H_{t,k}$ is the decoding state for any TM-augmented model $P(y_t|\mathbf{x}, \mathbf{y}_{<t}, \mathbf{z}_k)$ similar to $H_{t,\mathbf{Z}}$ in Eq.~\eqref{equation:TM-nmt-original}. 
For example, if we implement the weighted ensemble model on top of the model architecture of \citet{cai-etal-2021-neural}, then the source sentence and the $k$-th TM are encoded by two separate encoders respectively, resulting in a hidden state $H_t$ for current time step and a contextualized TM representation $H_{t,k}$ for the $k$-th TM in current time step. 


Similar to the average ensemble method, we do not train the whole network from scratch. Instead, we start from the trained model parameters of $P(y_t|\mathbf{x}, \mathbf{y}_{<t}, \mathbf{z}_k)$ in \S\ref{subsection:change_number_of_TMs} and then we just fine-tune the whole parameters including those from both $P(y_t|\mathbf{x}, \mathbf{y}_{<t}, \mathbf{z}_k)$ and $w(\mathbf{x}, \mathbf{y}_{<t}, \mathbf{z}_{k})$ for only about 2,000 updates on $90\%$ part of valid data while the other $10\%$ part of valid data are used to select the checkpoint for testing.

\paragraph{Remark} The above average ensemble model involves $K$ retrieved sentence pairs similar to the standard translation memory augmented models~\cite{gu2018search,zhang-etal-2018-guiding,cai-etal-2021-neural}, but the notable difference is that our model $P(y_t | \mathbf{x}, \mathbf{y}_{<t}, \mathbf{z}_{k})$ conditions on a single $\mathbf{z}_{k}$ whereas those standard TM-augmented NMT models (see Eq.~\eqref{equation:TM-nmt-original}) directly condition on $K$ retrieved sentence pairs $\mathbf{Z}=\{\mathbf{z}_k\}_{k=1}^K$.  


\subsection{Empirical Analysis on Bias and Variance}
\label{subsection:empirical_analysis_variance}

To verify the effectiveness of the proposed methods empirically, we conduct the variance-bias decomposition experiments similar to \S\ref{subsection:bias_variance_decomposition}. 
Specifically, we estimate the variance and bias of both three proposed models and two baselines (please refer to \S\ref{subsection:settings} for detailed settings).

Table ~\ref{table:variance_all_models} shows the overall results of estimated variance and bias.
By comparing results of different models, we can get the following two observations:
Firstly, through the results of variance, we can find that the three proposed methods all achieve lower variance compared with the default TM-augmented NMT model. It is notable that the weighted ensemble method achieves the best variance within all the TM-augmented NMT models.
Secondly, through the results of bias, we can find that all TM-augmented NMT models achieve lower bias compared to vanilla Transformer~\citep{vaswani2017attention} without TM. Although our methods have a slightly higher bias, the proposed weighted ensemble method can achieve comparable bias with regard to the default TM-augmented NMT model.


\begin{table}[t]
\small
\renewcommand\arraystretch{2}
\centering
\begin{tabular}{m{0.8cm}<{\centering}|m{1.0cm}<{\centering}|m{0.8cm}<{\centering}|m{0.8cm}<{\centering}|m{0.8cm}<{\centering}|m{0.8cm}<{\centering}}
\hline
{\textbf{Model}} &  \makecell[c]{w/o TM} & {TM-base} &  {TM-single} & {TM-average} & {TM-weight} \\
\hline
{\textbf{Var}} &  0.1573 & 0.2168 & 0.1944  &  0.1918  & \textbf{0.1814} \\
\hline
{\textbf{$\text{Bias}^2$}} &  1.9992 & \textbf{1.8460} & 1.9369  &  1.9395  & 1.9137 \\
\hline
\end{tabular}
\caption{Estimated variance and bias. }
\label{table:variance_all_models}
\end{table}

\section{Experiments}
\label{section:experiments}
In this section, we validate the effectiveness of the proposed methods in three scenarios: (1) the low-resource scenario where training pairs are scarce, (2) the plug-and-play scenario where additional bilingual pairs are added as the data store and the model is not re-trained any more as the data store is enlarged, and (3) the conventional high-resource scenario where the entire training data are used for training and retrieval. 
We use BLEU score \citep{papineni-etal-2002-bleu} as the automatic metric for the translation quality evaluation.

\subsection{Settings}
\label{subsection:settings}
\paragraph{Data} 
We use the JRC-Acquis corpus \citep{steinberger-etal-2006-jrc} and select four translation directions including Spanish$\Rightarrow$English (Es$\Rightarrow$En), En$\Rightarrow$Es, German$\Rightarrow$English (De$\Rightarrow$En), and En$\Rightarrow$De, for evaluation. Besides, we also use the re-split version of the Multi-Domain data set in \citet{aharoni-goldberg-2020-unsupervised} originally collected by \citet{koehn-knowles-2017-six} for our experiments, which includes five domains: Medical, Law, IT, Koran and Subtitle. Detailed data statistics and descriptions are shown in Appendix \ref{appendix_sec:data_statistics}.

\paragraph{Models} 
To study the effect of the proposed methods in \S\ref{section:proposed_approach}, we implement a series of model variants by using the fairseq toolkit~\citep{ott2019fairseq}. \#1 vanilla NMT without TM \citep{vaswani2017attention}. We remove the model components related to TM, and only employ the encoder-decoder architecture for NMT. \#2 Default TM-augmented NMT with top-5 TMs. We use top-5 TMs to train and test the model. Note that this is also a baseline model in \citet{cai-etal-2021-neural}. \#3 TM-augmented NMT with single TM. To study the effect of technique 1 in \S\ref{subsection:change_number_of_TMs} which conditions only one TM, we use top-1 TM to test the model. In order to avoid overfitting, during each epoch we use training pairs with top-5 TMs and empty TM to train the model six times, which is similar to \citet{he-etal-2021-fast}. \#4 TM-augmented NMT with the average ensemble. To study the effect of technique 2 in \S\ref{subsection:change_number_of_TMs} which enlarges the sample size, we use the trained model in \#3 directly and average ensemble top-5 TMs. \#5 TM-augmented NMT with the weighted ensemble. We fine-tune the trained model in \#3 with weighted ensemble top-5 TMs. Detailed model descriptions and experimental settings are in Appendix \ref{appendix_sec:backbone} and \ref{appendix_sec:detailed_training_settings} respectively.

\subsection{Low-Resource Scenario}
\label{subsection:low_resource_scenarios}

\begin{table*}[t]
\centering
\begin{tabular}{l|c|c|c|c|c|c}
\hline
\textbf{Model} & \textbf{Medical}  &  \textbf{Law}  &  \textbf{IT}  &  \textbf{Koran}  &  \textbf{Subtitle} & \textbf{Average} \\
\hline
\makecell[l]{\#1 w/o TM}~\citep{vaswani2017attention}& 47.62 & 50.85 & 34.40 & 14.45 & 20.22 & 33.51 \\
\hline
\makecell[l]{\#2 TM-base}~\citep{cai-etal-2021-neural} & 43.53 & 49.36 & 32.76 & 14.43 & 20.03 & 32.02 \\
\hline
\makecell[l]{\#3 TM-single  (ours \S\ref{subsection:change_number_of_TMs} Technique 1) }  & 47.04 & 50.84 & 35.33 & 14.80 & 21.22 & 33.85 \\
\hline
\makecell[l]{\#4 TM-average (ours \S\ref{subsection:change_number_of_TMs} Technique 2) }    & 47.13 & 50.82 & 35.50 & 14.87 & 21.30 & 33.92 \\
\hline
\makecell[l]{\#5 TM-weight (ours \S\ref{sub:weighted})}   & \textbf{47.97} & \textbf{52.28} & \textbf{35.84} & \textbf{16.59} & \textbf{22.58} & \textbf{35.05} \\
\hline
\end{tabular}
\caption{Experimental results (test set BLEU scores) on Multi-Domain dataset under low-resource scenario.}
\label{table:results_low_resource_multi_domain}
\end{table*}

One of the major advantages of our proposed weighted ensemble is that it has a lower variance, which means that it is less sensitive to fluctuations in the training data.
This motivates us to conduct experiments in low-resource scenarios, where we use only a part of the training data to train models. Specifically, we create low-resource scenario by randomly partitioning each training set in JRC-Acquis corpus and Multi-Domain dataset into four subsets of equal size. 
Then we only use the training pairs in the first subset to train each model.

\paragraph{Results}
The test results of the above models on Multi-Domain dataset and JRC-Acquis corpus are presented in Table ~\ref{table:results_low_resource_multi_domain} and Table ~\ref{table:results_low_resource_jrc} respectively.
We can get the following observations: 
(1) Our proposed weighted ensemble method delivers the best performance on test sets across all translation tasks, outperforming the vanilla Transformer by a large margin.
(2) The performance of default TM-augmented NMT with top-5 TMs is degraded compared with vanilla Transformer, while single TM method and average ensemble method make up for the degradation issue to some extent.

\begin{table}[t]
\centering
\small
\begin{tabular}{l|c|c|c|c}
\hline
\textbf{Model} & \textbf{Es$\Rightarrow$En}  & \textbf{En$\Rightarrow$Es} & \textbf{De$\Rightarrow$En} & \textbf{En$\Rightarrow$De} \\
\hline
\makecell[l]{w/o TM} & 58.44  & 56.11  &  54.54  &  49.97   \\
\hline
\makecell[l]{TM-base} & 57.31  & 55.06  &  53.92  &   48.67  \\
\hline
\makecell[l]{TM-single} & 57.56  & 55.24  &  54.03  &   48.82 \\
\hline
\makecell[l]{TM-average} & 57.08  & 54.91  &  53.77  &   48.41 \\
\hline
\makecell[l]{TM-weight} & \textbf{59.14}  & \textbf{56.53}  &  \textbf{55.36}  &  \textbf{50.51}  \\
\hline
\end{tabular}
\caption{Experimental results (test set BLEU scores) on four translation tasks of JRC-Acquis corpus under low-resource scenario.}
\label{table:results_low_resource_jrc}
\end{table}

\subsection{Plug-and-Play Scenario}
“Plug-and-play” is one of the most remarkable properties of TM \citep{cai-etal-2021-neural}, that is to say, the corpus used for TM retrieval is different during training and testing. This is useful especially in online products because we can adapt a trained model to a new corpus without retraining by adding or using a new TM. Specifically, we directly use models trained in \S \ref{subsection:low_resource_scenarios}, and add the second, the third, and the last subset to the TM data store gradually. At each datastore size, we retrieve for the test set again and test the performance of models with the newly retrieved TMs.

\begin{figure*}[!htp]
    \centering
    \includegraphics[width=\textwidth]{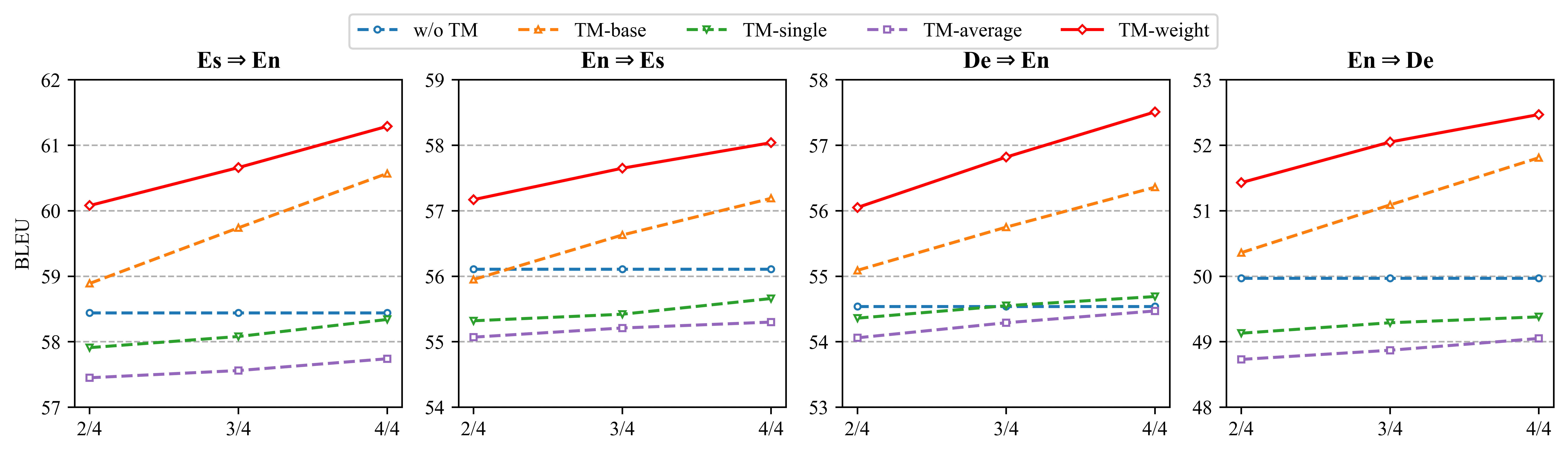} \\
    \caption{Test results of JRC-Acquis corpus under plug and play scenario with 2/4, 3/4 and 4/4 TMs respectively.}
    \label{fig:plug_play_jrc}
\end{figure*}

\begin{figure*}[!htp]
    \centering
    \includegraphics[width=\textwidth]{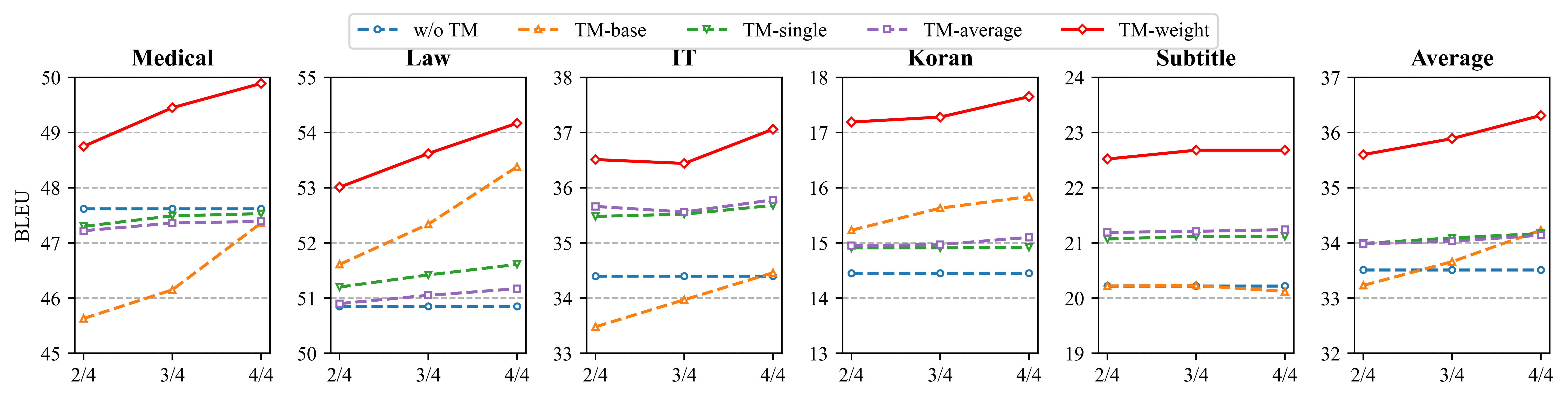} \\
    \caption{Test results of Multi-Domain dataset under plug and play scenario with 2/4, 3/4 and 4/4 TMs respectively.}
    \label{fig:plug_play_multi_domain}
\end{figure*}

\paragraph{Results}
Figure ~\ref{fig:plug_play_jrc} and Figure ~\ref{fig:plug_play_multi_domain} show the main results on the test sets of JRC-Acquis corpus and Multi-Domain dataset respectively. The general patterns are consistent across all experiments: The larger the TM becomes, the better translation performance the model achieves. When using all training data (4/4), the translation quality is boosted significantly. At the same time, our proposed weighted ensemble method achieves the best performance all the time. 

\subsection{High-Resource Scenario}
Following prior works \citep{he-etal-2021-fast, cai-etal-2021-neural} in TM-augmented NMT, we also conduct experiments in the high-resource scenario where all the training pairs of JRC-Acquis corpus are used to train models in order to prove the practicability and universality of our proposed methods.

\paragraph{Results}
The results  are presented in Table ~\ref{table:results_conventional_jrc}. We can see that our proposed weighted ensemble method still has the best performance even under the high-resource scenario, where default TM-augmented NMT using top-5 TMs is also relatively good because in this condition the bias plays an important role in fitting the whole training data to achieve good performance.

\begin{table}[t]
\centering
\small
\begin{tabular}{l|c|c|c|c}
\hline
\textbf{Model}  &  \textbf{Es$\Rightarrow$En}  & \textbf{En$\Rightarrow$Es} & \textbf{De$\Rightarrow$En} & \textbf{En$\Rightarrow$De}  \\
\hline
\makecell[l]{w/o TM} & 63.26  & 61.63  &  60.83  &  54.95   \\
\hline
\makecell[l]{TM-base} & 66.42  & 62.81  &  63.76  &   57.79  \\
\hline
\makecell[l]{TM-single} & 65.69  & 62.59  &  63.34  &   57.40 \\
\hline
\makecell[l]{TM-average} & 65.29  & 62.72  &  63.01  &   57.42 \\
\hline
\makecell[l]{TM-weight} & \textbf{66.89}  & \textbf{63.61}  &  \textbf{64.29}  &  \textbf{58.67}  \\
\hline
\end{tabular}
\caption{Test results on four translation tasks of JRC-Acquis corpus under high-resource scenario.}
\label{table:results_conventional_jrc}
\end{table}

\section{Related Work}

This work mainly contributes to the research line of Translation Memory (TM) augmented Neural Machine Translation (NMT).
Early works mainly concentrated on model architecture designing \citep{feng-etal-2017-memory, gu2018search, cao-xiong-2018-encoding, DBLP:conf/ecai/CaoKX20, zhang-etal-2018-guiding, Xia_Huang_Liu_Shi_2019, bulte-tezcan-2019-neural, xu-etal-2020-boosting, he-etal-2021-fast, wang-etal-2022-training, hoang2022interactions,cai2022recent}. 
Recently, \citet{cai-etal-2021-neural} used a dense retrieval method with a dual encoder to compute the similarity between the source sentence and TMs, which can use monolingual translation memory instead of bilingual ones. 
\citet{hoang2022improve_Robustness_TM_NMT} proposed to shuffle the retrieved suggestions to improve the robustness of TM-augmented NMT models.
\citet{2022chengNMTwithContrastiveTM} proposed to contrastively retrieve translation memories that are holistically similar to the source sentence while individually contrastive to each other providing maximal information gains.


The distinctions between our work and prior works are obvious: (1) We rethink TM-augmented NMT from a probabilistic view of retrieval and the variance-bias decomposition principle;
(2) We consider the performance of TM-augmented NMT under low-resource scenario, plug-and-play scenario, and conventional high-resource scenario, instead of only high-resource scenario.
(3) Our methods are agnostic to specific model and retrieval method and thus can be applied on top of any advanced architecture.


Another research line highly related to our work is $k$NN-MT ~\citep{Khandelwal2021knnmt, zheng-etal-2021-adaptive,zheng-etal-2021-non-parametric, meng-etal-2022-fast, 2021wangFasterKNNMT, wang-etal-2022-efficient, 2022zhuWhatKnowledge_ExplainableMemory, wang-etal-2022-learning-decoupled, du2022knnspeech, du2023federated, Wang_Wei_Zhang_Huang_Xie_Chen_2022knnHFNMT}.  However, $k$NN-MT retrieves similar key-value pairs on the token level, resulting in slow retrieval speed \citep{meng-etal-2022-fast, dai2023simple} and large storage space \citep{wang-etal-2022-efficient, 2022zhuWhatKnowledge_ExplainableMemory}, whereas TM-augmented NMT retrieves similar pairs on the sentence level, which has more efficient retrieval speed and takes up less storage space.


\section{Conclusion}
Existing work surprisingly finds that TM-augmented NMT fails to advance the NMT model under low-resource scenario, but it is unclear the reason why such a contradictory phenomenon happens.
This paper rethinks TM-augmented NMT from latent variable probabilistic model view and variance-bias decomposition view respectively, giving the reason for the failure under low-resource scenario.
Estimation for variance and bias indicates that TM-augmented NMT is better at fitting data (bias) yet worse at sensitivity to fluctuations in the training data (variance).
To better trade off the bias and variance, this paper proposes a simple yet effective weighted ensemble method for TM-augmented NMT.
Experiments under three scenarios demonstrate that the proposed method outperforms both vanilla Transformer and baseline TM-augmented models consistently. Future work could aim to find factors that influence the bias of TM-augmented NMT models, such as the quality of retrieved TMs.

\section*{Limitations}

Compared with the standard NMT, TM-augmented NMT models incur extra retrieval time during both training and inference. 
Besides, since both TMs and the source sentence need to be encoded, the speed of TM-augmented NMT is slower than that of standard NMT.
These characteristics indeed slightly induce some overhead. Specifically, experiments show that the latency time cost for all TM-augmented NMT models is higher than for vanilla Transformers. And the decoding time of our proposed weighted ensemble method is twice as much as that of the standard NMT.
This issue is one limitation of TM-augmented NMT, which can be further studied and addressed.

\section*{Ethics Statement}
This paper will not pose any ethical problems. First, machine translation is a standard task in natural language processing. Second, the datasets used in this paper have been already used in previous papers.

\section*{Acknowledgements}
Hongkun and Rui are with MT-Lab, Department of Computer Science and Engineering,
School of Electronic Information and Electrical Engineering, and also with the MoE Key Lab of Artificial Intelligence, AI Institute, Shanghai Jiao Tong
University, Shanghai 200204, China. Rui is supported by the General Program of National Natural Science Foundation of China (6217020129), Shanghai
Pujiang Program (21PJ1406800), Shanghai
Municipal Science and Technology Major Project
(2021SHZDZX0102), Beijing Academy of Artificial Intelligence (BAAI) (No. 4), CCF-Baidu Open Fund (F2022018), and the Alibaba-AIR Program (22088682). 

\normalem
\bibliography{anthology,custom}
\bibliographystyle{acl_natbib}

\appendix

\section{Proof of Proposition ~\ref{proposition:variance} in \S\ref{section:proposed_approach}}
\label{appendix_sec:proof_of_lemma}

Throughout this section, we prove the Proposition ~\ref{proposition:variance} in \S\ref{section:proposed_approach}. For simplicity, we use $f(\mathbf{z})$ and $f(\mathbf{z}_k)$ to denote $P(y_t | \mathbf{x}, \mathbf{y}_{<t}, \mathbf{z})$ and $P(y_t | \mathbf{x}, \mathbf{y}_{<t}, \mathbf{z}_k)$ in Proposition ~\ref{proposition:variance}, respectively. Note that each $\mathbf{z}_k$ are i.i.d sampled from $P(\mathbf{z}|\mathbf{x})$. 

\begin{proof}

Firstly, for the relationship between $\mathbb{E}_\mathbf{z} f(\mathbf{z})$ and $\mathbb{E}_{\mathbf{z}_1, \cdots, \mathbf{z}_K} \Big( \frac{1}{K} \sum_k   f(\mathbf{z}_k) \Big)$ we can get the following equations:
\begin{equation*}
\begin{aligned}
      & \quad\; \mathbb{E}_{\mathbf{z}_1, \cdots, \mathbf{z}_K} \Big( \frac{1}{K} \sum_k   f(\mathbf{z}_k) \Big) \\
      & = \frac{1}{K} \mathbb{E}_{\mathbf{z}_1, \cdots, \mathbf{z}_K} \Big(\sum_k   f(\mathbf{z}_k) \Big)   \\
      & =  \frac{1}{K} \sum_k \Big( \mathbb{E}_{\mathbf{z}_k}  f(\mathbf{z}_k) \Big)   \\
      & =  \frac{1}{K} \sum_k \Big( \mathbb{E}_{\mathbf{z}}  f(\mathbf{z}) \Big)   \\
      & =  \frac{1}{K} * K * \mathbb{E}_{\mathbf{z}}  f(\mathbf{z})   \\
      & =  \mathbb{E}_\mathbf{z} f(\mathbf{z}) . 
\end{aligned}
\end{equation*}

Secondly, we start from the relationship between $\mathbb{V}_\mathbf{z} f(\mathbf{z})$ and $\mathbb{V}_{\mathbf{z}_1, \mathbf{z}_2} \big( f(\mathbf{z}_1) + f(\mathbf{z}_2) \big)$:
\begin{equation*}
\begin{aligned}
      & \quad\; \mathbb{V}_{\mathbf{z}_1,\mathbf{z}_2} \big( f(\mathbf{z}_1) + f(\mathbf{z}_2) \big)  \\
      & = \mathbb{E}_{\mathbf{z}_1, \mathbf{z}_2} \Big(f(\mathbf{z}_1) + f(\mathbf{z}_2)  \\
      & \quad\; - \mathbb{E}_{\mathbf{z}_1,\mathbf{z}_2} \big(f(\mathbf{z}_1) + f(\mathbf{z}_2) \big) \Big)^2   \\
      & =  \mathbb{E}_{\mathbf{z}_1, \mathbf{z}_2} \big( f(\mathbf{z}_1) + f(\mathbf{z}_2) \big)^2 \\
      & \quad\; - \Big(\mathbb{E}_{\mathbf{z}_1,\mathbf{z}_2} \big( f(\mathbf{z}_1) + f(\mathbf{z}_2) \big) \Big)^2   \\
      & =  \mathbb{E}_{\mathbf{z}_1} \big( f(\mathbf{z}_1) \big)^2 + \mathbb{E}_{\mathbf{z}_2} \big( f(\mathbf{z}_2) \big)^2 \\
      & \quad\; + 2 \mathbb{E}_{\mathbf{z}_1,\mathbf{z}_2} \big( f(\mathbf{z}_1)f(\mathbf{z}_2) \big) \\ 
      & \quad\; - \big( \mathbb{E}_{\mathbf{z}_1} f(\mathbf{z}_1) \big)^2 - \big( \mathbb{E}_{\mathbf{z}_2} f(\mathbf{z}_2) \big)^2   \\
      & \quad\; - 2 \big( \mathbb{E}_{\mathbf{z}_1} f(\mathbf{z}_1) \big) \big( \mathbb{E}_{\mathbf{z}_2}f(\mathbf{z}_2) \big)  \\
      & =  \mathbb{E}_{\mathbf{z}_1} \big( f(\mathbf{z}_1) \big)^2 + \mathbb{E}_{\mathbf{z}_2} \big( f(\mathbf{z}_2) \big)^2 \\
      & \quad\; + 2 \big( \mathbb{E}_{\mathbf{z}_1} f(\mathbf{z}_1) \big) \big( \mathbb{E}_{\mathbf{z}_2}f(\mathbf{z}_2) \big) \\ 
      & \quad\; - \big( \mathbb{E}_{\mathbf{z}_1} f(\mathbf{z}_1) \big)^2 - \big( \mathbb{E}_{\mathbf{z}_2} f(\mathbf{z}_2) \big)^2   \\
      & \quad\; - 2 \big( \mathbb{E}_{\mathbf{z}_1} f(\mathbf{z}_1) \big) \big( \mathbb{E}_{\mathbf{z}_2}f(\mathbf{z}_2) \big)  \\
      & =  \mathbb{E}_{\mathbf{z}_1} \big( f(\mathbf{z}_1) \big)^2 - 
     \big(\mathbb{E}_{\mathbf{z}_1} f(\mathbf{z}_1)\big)^2  \\
     & \quad\; + \mathbb{E}_{\mathbf{z}_2} \big( f(\mathbf{z}_2) \big)^2 - 
     \big(\mathbb{E}_{\mathbf{z}_2} f(\mathbf{z}_2)\big)^2   \\
      & =  \mathbb{V}_{\mathbf{z}_1}  f(\mathbf{z}_1) + \mathbb{V}_{\mathbf{z}_2}  f(\mathbf{z}_2)    \\
      & = 2 \mathbb{V}_{\mathbf{z}}  f(\mathbf{z}).
\end{aligned}
\end{equation*}

Then, we can get the relationship between $\mathbb{V}_\mathbf{z} f(\mathbf{z})$ and $\mathbb{V}_{\mathbf{z}_1, \mathbf{z}_2} \Big( \frac{1}{2} \big( f(\mathbf{z}_1) + f(\mathbf{z}_2) \big) \Big)$ as follows:
\begin{equation*}
\begin{aligned}
      & \quad\; \mathbb{V}_{\mathbf{z}_1,\mathbf{z}_2} \Big( \frac{1}{2} \big( f(\mathbf{z}_1) + f(\mathbf{z}_2) \big) \Big)  \\
      & = \frac{1}{2^2} \mathbb{V}_{\mathbf{z}_1, \mathbf{z}_2}   \big( f(\mathbf{z}_1) + f(\mathbf{z}_2) \big)    \\
      & = \frac{1}{4} * 2 * \mathbb{V}_{\mathbf{z}}  f(\mathbf{z})   \\
      & =  \frac{1}{2} \mathbb{V}_\mathbf{z} f(\mathbf{z}) \\
      & <  \mathbb{V}_\mathbf{z} f(\mathbf{z}).
\end{aligned}
\end{equation*}

Similarly, for the relationship between $\mathbb{V}_\mathbf{z} f(\mathbf{z})$ and $\mathbb{V}_{\mathbf{z}_1, \cdots, \mathbf{z}_K} \Big( \frac{1}{K} \sum_k   f(\mathbf{z}_k) \Big)$ we can get the following equations:
\begin{equation*}
\begin{aligned}
      & \quad\; \mathbb{V}_{\mathbf{z}_1, \cdots, \mathbf{z}_K} \Big( \frac{1}{K} \sum_k   f(\mathbf{z}_k) \Big)  \\
      & = \frac{1}{K^2} \mathbb{V}_{\mathbf{z}_1, \cdots, \mathbf{z}_K} \Big(\sum_k   f(\mathbf{z}_k) \Big)   \\
      & = \frac{1}{K^2} \bigg( \sum_k \Big( \mathbb{E}_{\mathbf{z}_k} \big( f(\mathbf{z}_k) \big)^2 - 
     \big(\mathbb{E}_{\mathbf{z}_k} f(\mathbf{z}_k)\big)^2  \Big) \bigg) \\
      & = \frac{1}{K^2} \sum_k \Big( \mathbb{V}_{\mathbf{z}_k}  f(\mathbf{z}_k) \Big)   \\
      & = \frac{1}{K^2} \sum_k \Big( \mathbb{V}_{\mathbf{z}}  f(\mathbf{z}) \Big)   \\
      & = \frac{1}{K^2} * K * \mathbb{V}_{\mathbf{z}}  f(\mathbf{z})   \\
      & = \frac{1}{K} \mathbb{V}_\mathbf{z} f(\mathbf{z}) \\
      & \le  \mathbb{V}_\mathbf{z} f(\mathbf{z}) , 
\end{aligned}
\end{equation*}
where the equality holds if and only if $K = 1$.
\end{proof}

\section{Variance-Bias Estimation Method}
\label{appendix_sec:variance_estimation_algorithm}
In \S\ref{subsection:bias_variance_decomposition}, we mention that we follow classic methods \citep{hastie2009elements, yang2020rethinking} to estimate the variance and bias in Eq. \eqref{equation:decomposition}. Here we detail the method to estimate the variance, which is shown in Algorithm \ref{algorithm:estimate_variance}. Then $\text{Bias}^2$ is estimated by subtracting the variance from the loss \citep{yang2020rethinking}.
Specifically, in this paper we set $N$ to be 1 and $k$ to be 4 in Algorithm \ref{algorithm:estimate_variance}.
Besides, we compute the variance for each test point $(\mathbf{x}, \mathbf{y}_{<t})$ in test set, and average them to get the final variance for each model.

    \begin{algorithm}[htp]
    \renewcommand{\algorithmicrequire}{\textbf{Input:}}
    \renewcommand{\algorithmicensure}{\textbf{Output:}}
    \caption{Estimating Variance of NMT Models}
    \label{algorithm:estimate_variance}
    \begin{algorithmic}[1]
        \REQUIRE Test point $(\mathbf{x}, \mathbf{y}_{<t})$, Training data $\boldsymbol{\tau}$.
	\FOR{$i = 1$ to $k$}
            \STATE Split the $\boldsymbol{\tau}$ into $\boldsymbol{\tau}_1^{(i)}, \cdots , \boldsymbol{\tau}_N^{(i)}$.   
            \FOR{$j = 1$ to $N$}
                \STATE Train the model using $\boldsymbol{\tau}_j^{(i)}$;
                \STATE Evaluate the model at $(\mathbf{x}, \mathbf{y}_{<t})$; call the result $P_j^{(i)}(y|\mathbf{x}, \mathbf{y}_{<t})$;
                \STATE Normalize the top 100 probability in $P_j^{(i)}(y|\mathbf{x}, \mathbf{y}_{<t})$ and setting others to be 0 \\
                (to reduce computation complexity);
            \ENDFOR
        \ENDFOR
        \STATE Compute \quad\quad $\hat{P}(y|\mathbf{x}, \mathbf{y}_{<t}) = $ \\
        $\exp\big( \frac{1}{N \cdot k} \sum_{i,j} \log P_j^{(i)}(y|\mathbf{x}, \mathbf{y}_{<t}) \big)$ \\
        ($\hat{P}(y|\mathbf{x}, \mathbf{y}_{<t})$ estimates $\overline{P}(y|\mathbf{x}, \mathbf{y}_{<t})$).
        \STATE Normalize $\hat{P}(y|\mathbf{x}, \mathbf{y}_{<t})$ to get a probability distribution.
        \STATE Compute the variance \quad\quad $var = $  \\
        $\frac{1}{N \cdot k} \sum_{ij} D_{KL} \big( P_j^{(i)}(y|\mathbf{x}, \mathbf{y}_{<t}) || \hat{P}(y|\mathbf{x}, \mathbf{y}_{<t}) \big)$	
	\ENSURE  $var$
    \end{algorithmic}  
\end{algorithm}

\section{Details about Architecture}
\label{appendix_sec:backbone}
In \S\ref{subsection:bias_variance_decomposition}, we use single encoder backbone \citep{bulte-tezcan-2019-neural} and dual encoder backbone \citep{cai-etal-2021-neural} respectively. Here we provide a detailed description of these two architectures.

\paragraph{Single Encoder Architecture}
For this architecture, we follow \citet{bulte-tezcan-2019-neural}. Specifically, the model architecture is the same as vanilla Transformer, which consists of an encoder and a decoder. The encoder encodes the concatenation of the source sentence and TM, relying on the encoder's self-attention to compare the source sentence to the TM and determine which TM phrases are relevant for the translation\citep{hoang2022interactions}. Therefore, the only difference is that we need to change the input format to the concatenation of the source sentence and TM.

\paragraph{Dual Encoder Architecture}
This architecture is similar to that used by \citet{cai-etal-2021-neural}, which achieves relatively good performance and thus can serve as a strong baseline~\citep{cai-etal-2021-neural}. 


On the encoder side, the source sentence and TM are encoded by two separate encoders respectively, resulting in a set of contextualized token representations $\{z_{kj}\}_{j=1}^{L_k}$, where $L_k$ is the length of the $k$-th TM.

On the decoder side, after using $\mathbf{y_{<t}}$ and $\mathbf{x}$ to get the hidden state $H_t$ at current time step $t$, it can produce the original NMT probability $P_{nmt}(y_t | \mathbf{x}, \mathbf{y}_{<t})$ and the TM probability $P_{tm}(y_t | \mathbf{x}, \mathbf{y}_{<t}, \mathbf{Z})$ concurrently, and finally get the probability $P(y_t | \mathbf{x}, \mathbf{y}_{<t}, \mathbf{Z})$.

Specifically, the original NMT probability $P_{nmt}(y_t | \mathbf{x}, \mathbf{y}_{<t})$ is defined by as follows:
\begin{equation}
\label{equation:P_nmt_in_TMNMT}
    P_{nmt}(y_t | \mathbf{x}, \mathbf{y}_{<t}) = \mathrm{Softmax} \big( f(H_t) \big)[y_t],
\end{equation}
where $f$ is a linear layer that maps hidden state $H_t$ to a probability distribution.
And for the TM probability $P_{tm}(y_t | \mathbf{x}, \mathbf{y}_{<t}, \mathbf{Z})$, the decoder firstly computes a cross attention score $\alpha = \{\alpha_{kj}\}$ over all tokens of the $k$-th TM sentence where $\alpha_{kj}$ is the attention score of $H_t$ to the $j$-th token in the $k$-th TM. The computation is shown as follows:
\begin{equation}
\label{equation:alpha_kj_in_TMNMT}
    \alpha_{kj} = \frac{\exp(H_t^T W_{tm} z_{kj})}{\sum_{k=1}^K \sum_{j=1}^{L_k} \exp(H_t^T W_{tm} z_{kj})} ,
\end{equation}
where $W_{tm}$ are parameters.
Then we can get the contextualized TM representation $H_{t,\mathbf{Z}}$ for the TM $\mathbf{Z}$ in current time step, as shown in Eq.~\eqref{equation:h_tk_in_TMNMT}:
\begin{equation}
\label{equation:h_tk_in_TMNMT}
    H_{t, \mathbf{Z}} = W_h \sum\limits_{k=1}^{L} \sum\limits_{j=1}^{L_k} \alpha_{kj}z_{kj},
\end{equation}
where $W_{h}$ are parameters.
Then, the TM probability $P_{tm}(y_t | \mathbf{x}, \mathbf{y}_{<t}, \mathbf{Z})$ can be computed as follows:
\begin{equation}
\label{equation:p_tm_in_TMNMT}
    P_{tm}(y_t | \mathbf{x}, \mathbf{y}_{<t}, \mathbf{Z}) = \sum\limits_{k=1}^{K}\sum\limits_{j=1}^{L_k}\alpha_{kj} \mathbb I_{z_{kj}=y_t}  ,
\end{equation}
where $\mathbb I$ is the indicator function.



Finally, the decoder computes the next-token probability as follows:
\begin{multline}
\label{equation:p_final_in_TMNMT}
    P(y_t | \mathbf{x}, \mathbf{y}_{<t}, \mathbf{Z})  = (1-\lambda_t)P_{nmt}(y_t | \mathbf{x}, \mathbf{y}_{<t}) + \\
    \lambda_t P_{tm}(y_t | \mathbf{x}, \mathbf{y}_{<t}, \mathbf{Z}) 
\end{multline}
where $\lambda_t$ is a gating variable computed by another linear layer whose input is $H_{t, \mathbf{Z}}$.





\section{Detailed Settings for All Experiments}
\label{appendix_sec:detailed_training_settings}

\begin{table*}[htpb]
    \centering
    \begin{tabular}{l | c c | c c}
    \toprule
       \multirow{2}{*}{\bf Model}  &  \multirow{2}{*}{\bf \# Param.} &  \multirow{2}{*}{\bf GPU Hours} & \multicolumn{2}{c}{\bf Hyperparam.}\\ 
       & & & learning rate & dropout \\
       \midrule
    Transformer   & 83M & 16h & 7e-4 & 0.1 \\
    TM-augmented Transformer   & 98M & 19h & 7e-4 & 0.1\\   
    \bottomrule
    \end{tabular}
    \caption{The number of parameters, training budget (in GPU hours), and hyperparameters of each model.}
    \label{table:training_budget}
\end{table*}

\begin{table*}[htpb]
    \centering
    \begin{threeparttable}
    \begin{tabular}{l l l}
    \toprule
         \bf Usage & \bf Package & \bf License  \\
         \midrule
         \multirow{2}{*}{Preprocessing} & mosesdecoder~\cite{koehn-etal-2007-moses}\tnote{1} &  LGPL-2.1 \\
         & subword-nmt~\cite{sennrich-etal-2016-neural}\tnote{2} & MIT \\
         \midrule
         Model training & fairseq~\cite{ott2019fairseq}\tnote{3} & MIT \\
         \midrule
         Evaluation &  BLEU~\cite{papineni-etal-2002-bleu}\tnote{1} & LGPL-2.1 \\
         \bottomrule   
    \end{tabular}
    \begin{tablenotes}
            \item[1] \url{https://github.com/moses-smt/mosesdecoder}\\
            \item[2] \url{https://github.com/rsennrich/subword-nmt} \\
            \item[3] \url{https://github.com/facebookresearch/fairseq}
        \end{tablenotes}
    \end{threeparttable}
    \caption{Packages we used for preprocessing, model training and evaluation.}
    \label{table:packages}
\end{table*}

Here we provide a detailed description of our configuration settings in the paper for the preliminary experiment (\S\ref{section:introduction}), the variance-bias decomposition experiment (\S\ref{subsection:bias_variance_decomposition} and \S\ref{section:proposed_approach}) and the main experiment (\S\ref{section:experiments}) respectively.

\paragraph{Settings for Preliminary Experiment in \S\ref{section:introduction}}
In this part, we implement vanilla Transformer base model \citep{vaswani2017attention} for the standard NMT. The learning rate schedule, dropout, and label smoothing are the same as \citet{vaswani2017attention}. For the TM-augmented NMT, we follow \citet{cai-etal-2021-neural} and implement the dual encoder architecture aforementioned. We conduct experiments on the JRC-Acquis German$\Rightarrow$English task. For high-resource scenario we use the full training data and train models with up to 100k steps. For low-resource scenario we randomly select a quarter of training data and train models with up to 30k steps.

\paragraph{Settings for Variance-Bias Decomposition Experiment in \S\ref{subsection:bias_variance_decomposition} and \S\ref{section:proposed_approach}}
For the experiment in \S\ref{subsection:bias_variance_decomposition}, we use both single encoder and dual encoder architecture introduced above in order to eliminate the effect of model architecture. 
For dual encoder, since the amount of parameters in vanilla Transformer is originally smaller than that of TM-augmented NMT model, which makes the two models non-comparable \citep{yang2020rethinking}, we use empty TMs to simulate vanilla Transformer and make the two models comparable. We set $N=1$ and $k=4$ in Algorithm \ref{algorithm:estimate_variance} to estimate the variance, so we train models with up to 30k steps. The other settings are the same as the preliminary experiment in \S\ref{section:introduction}.
For single encoder, the configuration is the same as vanilla Transformer base model \citep{vaswani2017attention}. And we also train models with up to 30k steps. For the experiment in \S\ref{section:proposed_approach}, we use dual encoder architecture as in Table \ref{table:variance_all_models}.

\paragraph{Settings for Main Experiment in \S\ref{section:experiments}}
We build our model using Transformer blocks with the same configuration as Transformer Base \citep{vaswani2017attention} (8 attention heads, 512 dimensional hidden state, and 2048 dimensional feed-forward state). 
The number of Transformer blocks is 4 for the memory encoder, 6 for the source sentence encoder, and 6 for the decoder side. 
We retrieve the top 5 TM sentences. The FAISS index code is "IVF1024\_HNSW32,SQ8" and the search depth is 64.

We follow the learning rate schedule, dropout, and label smoothing settings described in \citet{vaswani2017attention}. 
We use Adam optimizer \citep{kingma2014adam} and train models with up to 100K steps throughout all experiments.
When fine-tuning models with the proposed weighted ensemble method, we only need to use $90\%$ valid pairs to fine-tune with up to 2K steps and use the remaining $10\%$ valid pairs to choose the checkpoint for testing. 

Table \ref{table:training_budget} provides the number of parameters, training budget, and hyperparameters of each model. All experiments were performed on 8 V100 GPUs. We report the result of a single run for each experiment.

\paragraph{Packages}
Table \ref{table:packages} shows the packages we used for preprocessing, model training and evaluation.

\section{Details about Data Statistics}
\label{appendix_sec:data_statistics}
In this paper, we use the JRC-Acquis corpus \citep{steinberger-etal-2006-jrc} and the re-split version of the Multi-Domain data set in \citet{aharoni-goldberg-2020-unsupervised} for our experiments.

\paragraph{JRC-Acquis}
For the JRC-Acquis corpus, it contains the total body of European Union (EU) law applicable to the EU member states.
This corpus was also used by \citet{gu2018search, zhang-etal-2018-guiding, Xia_Huang_Liu_Shi_2019, cai-etal-2021-neural} and we managed to get the datasets originally preprocessed by \citet{gu2018search}. 
Specifically, we select four translation directions, namely, Spanish$\Rightarrow$English (Es$\Rightarrow$En), En$\Rightarrow$Es, German$\Rightarrow$English (De$\Rightarrow$En), and En$\Rightarrow$De, for evaluation.
Table ~\ref{table:statistics_jrc_dataset} shows the detailed number of train/dev/test pairs for each language pair.

\begin{table}[!htp]
\centering
\small
\begin{tabular}{c|c|c|c}
\hline
Dataset & \#Train Pairs & \#Dev Pairs & \#Test Pairs \\
\hline
\textbf{En $\Leftrightarrow$ Es} &  679,088  & 2,533  &  2,596  \\
\textbf{En $\Leftrightarrow$ Es} &  699,569  & 2,454  &  2,483  \\
\hline
\end{tabular}
\caption{Data statistics for the JRC-Acquis corpus.}
\label{table:statistics_jrc_dataset}
\end{table}

\paragraph{Multi-Domain Data Set}
For the Multi-Domain data set, it includes German-English parallel data in five domains: Medical, Law, IT, Koran, and Subtitle. 
Table ~\ref{table:statistics_multi_domain} shows the detailed number of train/dev/test pairs for each domain.

\begin{table}[!htp]
\centering
\small
\begin{tabular}{l|r|r|r}
\hline
Dataset & \#Train Pairs & \#Dev Pairs & \#Test Pairs \\
\hline
Medical & 245,553  & 2,000  &  2,000 \\
Law &  459,721  & 2,000  &  2,000  \\
IT & 220,241  & 2,000  &  2,000 \\
Koran & 17,833  & 2,000  &  2,000 \\
Subtitle & 499,969  & 2,000  &  2,000 \\
\hline
\end{tabular}
\caption{Data statistics for the multi-domain dataset.}
\label{table:statistics_multi_domain}
\end{table}

\end{CJK*}
\end{document}